\documentclass[sigconf]{acmart}

\AtBeginDocument{%
  \providecommand\BibTeX{{%
    \normalfont B\kern-0.5em{\scshape i\kern-0.25em b}\kern-0.8em\TeX}}}

\copyrightyear{2020}
\acmYear{2020}
\setcopyright{rightsretained}

\acmConference{VAM-HRI 2020}{23\textsuperscript{rd} Mar, 2020}{Cambridge, UK}
\acmDOI{}
\acmISBN{}

\usepackage{graphicx}
\usepackage{subcaption}
\usepackage{gensymb}
\usepackage{multirow}
\usepackage{pgf-pie}
\usepackage[]{quoting}

\begin{document}

\title{Towards an immersive user interface for waypoint navigation of mobile robots}

\author{Greg Baker}
\affiliation{%
  \institution{\textit{Bristol Robotics Laboratory} \\
  \textit{University of the West of England}}
  \city{Bristol}
  \country{UK}
}
\email{gregory.baker@brl.ac.uk}

\author{Tom Bridgwater}
\affiliation{%
  \institution{\textit{Bristol Robotics Laboratory} \\
  \textit{University of the West of England}}
  \city{Bristol}
  \country{UK}
}
\email{tom.bridgwater@uwe.ac.uk}

\author{Paul Bremner}
\affiliation{%
  \institution{\textit{Bristol Robotics Laboratory} \\
  \textit{University of the West of England}}
  \city{Bristol}
  \country{UK}
}
\email{paul.bremner@brl.ac.uk}

\author{Manuel Giuliani}
\affiliation{%
  \institution{\textit{Bristol Robotics Laboratory} \\
  \textit{University of the West of England}}
  \city{Bristol}
  \country{UK}
}
\email{manuel.giuliani@brl.ac.uk}

\renewcommand{\shortauthors}{Baker, Bridgwater, Bremner and Giuliani}


\begin{abstract}

In this paper, we investigate the utility of head-mounted display (HMD) interfaces for navigation of mobile robots. We focus on the selection of waypoint positions for the robot, whilst maintaining an egocentric view of the robot's environment. Inspired by virtual reality (VR) gaming, we propose a target selection method that uses the 6 degrees-of-freedom tracked controllers of a commercial VR headset. This allows an operator to point to the desired target position, in the vicinity of the robot, which the robot then autonomously navigates towards. A user study (37 participants) was conducted to examine the efficacy of this control strategy when compared to direct control, both with and without a communication delay. The results of the experiment showed that participants were able to learn how to use the novel system quickly, and the majority of participants reported a preference for waypoint control. Across all recorded metrics (task performance, operator workload and usability) the proposed waypoint control interface was not significantly affected by the communication delay, in contrast to direct control. The simulated experiment indicated that a real-world implementation of the proposed interface could be effective, but also highlighted the need to manage the negative effects of HMDs - particularly VR sickness.

\end{abstract}


\maketitle

\section{Introduction} \label{sec:introduction}

Autonomy is the capability of a system to carry out tasks without human control \cite{Beer2014TowardInteraction}. Research has shown that a systems with a higher level of autonomy can reduce operator workload and stress \cite{Vidulich2012MentalAwareness} and increase task performance \cite{sheridan1992telerobotics}, especially in the presence of communication delays between the operator and the the robot \cite{Chen2007HumanRobots,Luck2006AnLatency}. However, the complexity of the tasks and/or environment can limit the level of autonomy that can be achieved. Furthermore, a lower level of autonomy may be deemed preferable if human involvement in the exploration/navigation process is one of the primary functions of the robot (e.g. telepresence robots). Often, the optimum level of autonomy lies somewhere between the two extremes (fully manual/direct control and fully autonomous operation), necessitating at least some level of human-robot interaction (HRI).

A common theme among various implementations of robotic teleoperation is the desire to move the robot to a multitude of locations within the remote environment. This must be achieved in a manner which is both efficient and safe for the operator, the robot, and the environment. The user interface (UI), through which a mobile robot is remotely controlled and monitored, can significantly affect the speed and efficacy with which these exploration and navigation tasks can be achieved. Traditional UIs often use a joystick or keyboard as input devices, with visual feedback displayed on computer monitors \cite{Hainsworth2001TeleoperationRobotics}. Immersive interfaces, such as head-mounted displays (HMDs), provide an alternative interface for robot teleoperation. It has been shown that these types of interfaces can improve the operator’s situational awareness, and thus their ability to perform spatial tasks \cite{Garcia2017ARobots, Almeida2017ImprovingInterfaces, Jankowski2015UsabilityTeleoperation}. However, more work has to be done to explore the implications of immersive interfaces on human factors, and to ensure that they are suitable for real-world applications.

In this paper, we explore the implementation of waypoint navigation, a type of supervisory control, for a system using a HMD-based interface. In waypoint navigation, the operator selects target locations for the robot, and the robot moves to these locations autonomously. This method of control is regarded as a relatively low level of autonomy, because the operator is still responsible for all high-level mission and task planning; only local path-planning and movement is handled autonomously by the robot. For waypoint control, the UI must provide a means by which the operator can select target positions and/or orientations for the robot. Target-selection methods used in traditional UIs generally involve clicking a target location on a map (either 2D or 3D) \cite{Kay1995STRIPE:Data, Luck2006AnLatency, Schwarz2016SupervisedRobot, Goodrich2007ManagingExperiments}, or on an image (or stereoscopic pair of images) from the robot’s onboard camera \cite{Cameron1987FusingVehicles, Volpe1996ThePrototype, Mosiello2013UsingRobot, Bazzano2017ComparingTelepresence}, using a computer mouse or joystick as an input device. Modern, commerical HMD systems (e.g. HTC Vive and Oculus Rift), often provide 6 degrees-of-freedom (position and orientation) tracking for the headset and gaming controllers. The tracking of the controllers allows the user to easily and accurately point to features in the virtual environment (VE) - in a similar way to how one would use a laser pointer in the real world. This function is frequently used as an input method in VR gaming, owing to its intuitiveness and effectiveness. We have leveraged this functionality to design a pointing-based target selection method for robot teleoperation, inspired by similar methods used in VR games.

This paper first explores the literature related to waypoint navigation and HMD-based interfaces in mobile robotics. We then propose a target-selection method for HMDs, based on a method used in the VR gaming industry. We then describe a VR-based user study that we designed to test the usability and effectiveness of this target selection strategy, against a baseline condition of direct control. The results of the study are reported and the implications and limitations of these results are then discussed. We conclude with a summary of our findings and recommendations for future work.

\section{Background} \label{sec:background}

In this literature review, we focus on the user interfaces (UIs) through which waypoint navigation of mobile robots has been implemented. For a more general introduction to supervisory control methods we refer the reader to \citet{sheridan1992telerobotics}.

One of the main benefits of supervisory control is improved performance under communication delays between the operator and the robot. On top of this, supervisory control can also help reduce operator workload, because the operator is no longer responsible for the lowest level of control of the robot. While the robot autonomously handles a given task (under human supervision), the operator is able to perform other functions, such as surveying the remote environment using the robot’s cameras, improving their situational awareness \cite{Endsley1988DesignEnhancement}. According to \citet{Endsley1988DesignEnhancement}, increased situational awareness can improve task performance and reduce the chance of errors. However, supervisory control schemes can also negatively impact HRI, especially if the operator has a low level of trust in the system, or becomes bored in their supervisory role \cite{Cummings2013BoredomControl}. These effects have to be considered and managed to maximise the effectiveness of the system.

Waypoint navigation, as a form of supervisory control for mobile robots, has been around for over 30 years. Some of the early work in this area came from NASA’s Mars rover programme \cite{Wilcox1992RoboticExploration}. One of the challenges of designing a mobile robot for operation on Mars is the amount of time that it takes for a signal time to travel between Earth and Mars. \citet{Wilcox1992RoboticExploration} states that ‘it is impractical to have a rover that is teleoperated from Earth (that is, one in which the lowest-level feedback control is mediated through real-time perception of a human being)’. The paper mentions two proposed methods of implementing waypoint navigation for the rover; Computer-Aided Remote Driving (CARD) and semi-autonomous navigation (SAN). In CARD, the operator is presented with a pair of stereo images (using a stereoscopic display) from the robot’s onboard cameras, and selects a safe path for the robot using a 3D-cursor. In SAN, global routes for the rover are plotted by the operator using a topographic map of the Mars surface. Local routes are then adapted autonomously by the rover, based on more detailed information from the onboard sensors.

In order to enable the operator to select targets in 3D space for the robot to navigate to, two primary functions have to be achieved by the interface; (1) the robot’s environment has to be displayed to the operator, and (2) the operator has to be able to select a target point from that display using an input device. In terms of display, the environment is generally rendered to the operator in one of two ways; either using camera images transmitted from the robot (e.g. CARD) or using a 2D or 3D map (e.g. SAN).

Using an immersive display, such as an HMD, to render the robot’s environment, rather than a computer monitor, has been shown to improve the operator’s situational awareness and task performance in navigation tasks \cite{Garcia2017ARobots, Almeida2017ImprovingInterfaces, Roldan2019BringingInfrastructures}. Some of these benefits may stem from the ability to transform some explicit controls into implicit ones; for example controlling the orientation of the camera through movement of the user's head, rather than a joystick or other control \cite{Garcia2017ARobots}. Images from the robot can also be rendered across a much larger portion of the operator’s ocular field-of-view. Thus, objects in the remote environment that may appear quite small on a computer monitor, can appear much larger in a HMD. This means the operator simply has to shift their gaze to a region of interest in the rendered image, rather than manually zooming in and out of the image.

Despite these benefits, HMDs also pose additional challenges to interface design. In particular, the effects of ‘simulator sickness’ - namely nausea, headaches and/or vomiting - have been shown to be more severe when using HMDs, compared to a computer monitor display \cite{Schmidt2014UserInspection}. \citet{Moss2011CharacteristicsSickness} makes several recommendations to reduce these effects; (1) HMD update rate should be as high as possible (2) do not occlude the user’s peripheral visual of their local environment (if possible) (3) HMD users should be provided with postural support (e.g. a railing to grasp), rather than standing freely. Another consideration is that restricting, or completely removing the operators ability to see their local environment may inhibit their ability to cooperate with other operators, and may prevent the use of certain input devices. Finally, HMDs are not nearly as familiar to the majority of users as computer monitors, and their use can induce operator stress. These issues have so far limited the use of HMDs in real world applications. More research into these interfaces is necessary in order to capitalise on the aforementioned benefits, whilst minimising these negative impacts. 

Using HMD-based interfaces for waypoint navigation provides an interesting challenge. The way in which a target position is selected will be affected by the way in which the environment is being rendered to the operator. The camera images from the robot’s onboard camera(s) could be streamed directly to the HMD \cite{Oh2018360Tour, Almeida2017ImprovingInterfaces}. This technique restricts the viewpoint of the operator to the viewpoint of the robot’s camera, i.e. an egocentric viewpoint. In this case, waypoint selection methods would have to be augmented onto the camera images. Alternatively, the operator could be shown a 3D representation of the robot’s environment. This model may be produced from \textit{a priori} knowledge \cite{Wilson2018VETO:Tele-Operation, Roldan2019BringingInfrastructures}, or ‘virtualized’ in real-time using 3D mapping \cite{Kelly2011Real-timeControl, Schwarz2016SupervisedRobot, Tikanmaki2017TheReality, Stotko2019ARobot}. Rendering the environment in 3D allows the operator to separate their viewpoint from the robot’s position, allowing either egocentric, exocentric or tethered viewpoints \cite{Hollands2011ViewpointAwareness}. These two techniques can also be combined by projecting the camera images into the 3D environment, in the vein of the ecological interfaces developed by \citet{Nielsen2007EcologicalTeleoperation} (e.g. \cite{Varadarajan2011AugmentedRobots, Rodehutskors2015IntuitiveTracking, Schwarz2016SupervisedRobot, Allspaw2019DesignGames}), or by allowing the operator to switch between the two modes. Finally, in the situation where the operator and robot are collocated, the operator may view the environment directly, with the target selection method augmented on top of the scene. This could either be achieved using a dedicated augmented-reality HMD (e.g. Microsoft HoloLens) \cite{Walker2019Robot} or using video-passthrough with a VR-type HMD (e.g. HTC Vive, Oculus Rift); Class 3 and Class 4 interfaces respectively in \citeauthor{Milgram1994ADisplays}'s taxonomy \cite{Milgram1994ADisplays}.

The primary efforts to achieve waypoint selection with a HMD-based interface that we have found are \cite{Tikanmaki2017TheReality, Walker2019Robot, Roldan2019BringingInfrastructures, Allspaw2019DesignGames}. The target selection methods in these four implementations, which all use an exocentric viewpoint, are outlined below:

\subsubsection*{Drag and drop} \cite{Tikanmaki2017TheReality} - The user generates targets for the robot by dragging “ghost versions of the robot” to the desired locations in the virtually rendered 3D world, using the tracked VR headset (HTC Vive) controllers. To facilitate this target selection method, the operator is able to scale themselves up and down, as well as translate around the VE. The VE is pre-mapped, but updated in real-time by the teleoperated unmanned ground robot (UGV). A quantitative evaluation of the system is not provided in the paper.

\subsubsection*{Augmented reality virtual surrogate} \cite{Walker2019Robot} - The operator sets waypoints for a collocated unmanned aerial vehicle (UAV), by controlling a virtual ‘surrogate’ UAV superimposed on the environment using an augmented-reality HMD (Microsoft HoloLens). This control method was shown to improve navigation task completion time, as well as improve the operators ability to multitask during the trial, when compared to direct control. Although this implementation relies on collocation of the operator and the robot, the same waypoint setting method could be implemented remotely if the environment was rendered virtually (as in \cite{Tikanmaki2017TheReality}), or by superimposing the steerable ‘virtual leader’ robot on to the camera feed from the robot \cite{Sauer2010Mixed-RealityNetworks, Kelly2011Real-timeControl}.

\subsubsection*{Point and click} \cite{Roldan2019BringingInfrastructures,Allspaw2019DesignGames} - \citet{Roldan2019BringingInfrastructures} proposes a pointing-based target selection method, for the control of a UGV and UAV in an outdoor multi-robot navigation task. This target selection method is essentially the same as the one described in this paper, where the operator uses the 6DoF tracked controllers to point to a target location in the virtual world. The main difference between this system and ours is that \citet{Roldan2019BringingInfrastructures} use an exocentric viewpoint. A full 3D virtual copy of the real environment is modelled by hand from \textit{a priori} knowledge and presented to the operator using the HTC Vive headset. \citet{Allspaw2019DesignGames} describes a similar exocentric, pointing-based target selection method for the control of a humanoid robot. The authors also propose allowing the operator to place virtual footprint objects in the environment, that the robot would try to copy exactly.

As previously mentioned, exocentric viewpoints necessarily require the environment to be mapped in 3D. Therefore, robots that operate in unknown and unmapped environments would have a requirement for real-time 3D mapping. 3D maps produced by state-of-the-art real-time mapping techniques cannot replicate the level of detail of a camera image, and often require more specialist equipment. Alternatively, the camera feed from the robot's onboard camera could be streamed directly to the operator's HMD. This would allow the operator to perceive the environment in a high level of detail, but necessitates an egocentric viewpoint. Egocentric viewpoints have been shown to be preferable for local guidance \cite{Wickens2013EngineeringPerformance, Hollands2011ViewpointAwareness} and increased operator presence - which may make it a preferable for certain applications, such as telepresence. To this end, we propose an interface that combines pointing-based target selection with an egocentric viewpoint, in the following section. We also conducted a user study to test the efficacy of the proposed system. To the best of our knowledge, the effectiveness of interfaces that use a pointing-based target selection method (as per \cite{Roldan2019BringingInfrastructures} and \cite{Allspaw2019DesignGames}), with an egocentric viewpoint have not been assessed.

\section{System Design} \label{sec:system_design}

\begin{figure*}
  \centering
  \begin{subfigure}[b]{0.32\hsize}
    \centering
    \includegraphics[width=\hsize]{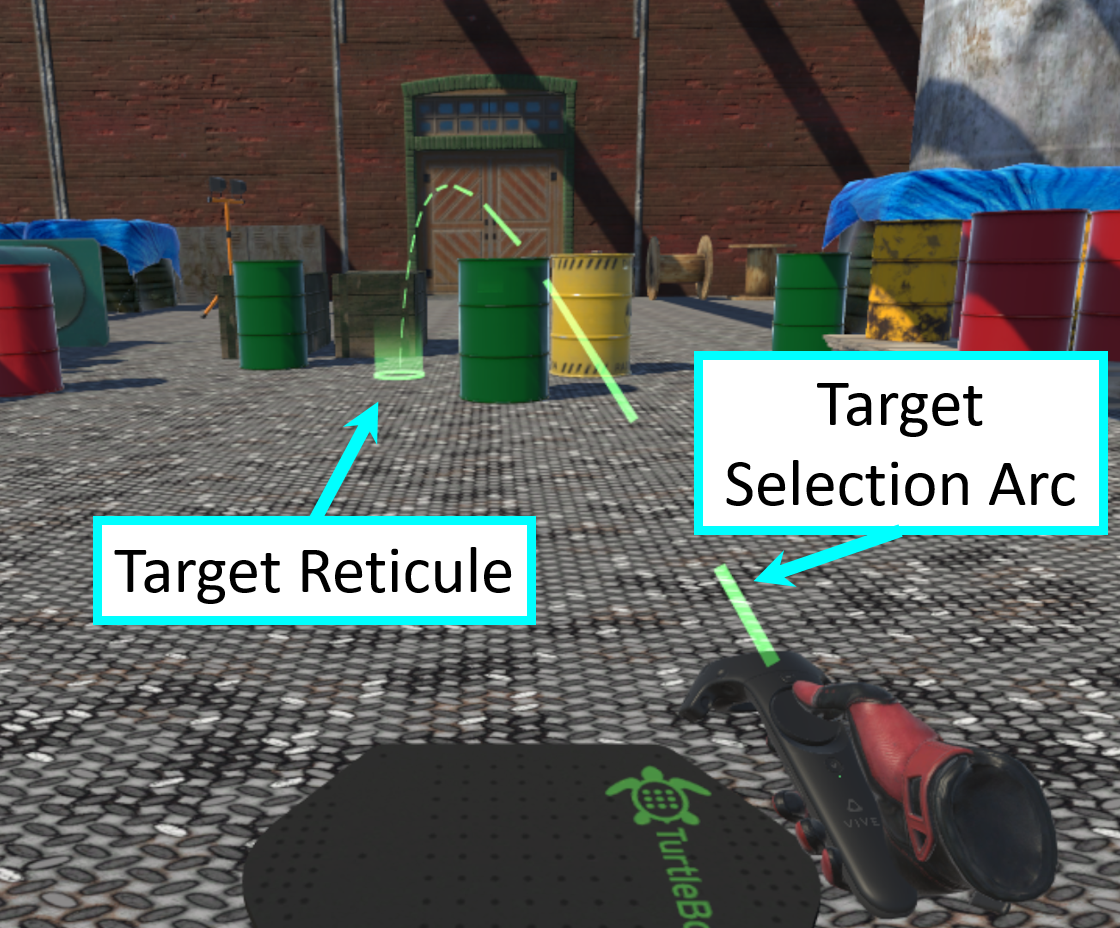} 
    \caption{Target selection method} 
    \label{fig:target_select}
  \end{subfigure}
  \hspace{0.2cm}
  \begin{subfigure}[b]{0.32\hsize}
    \centering
    \includegraphics[width=\hsize]{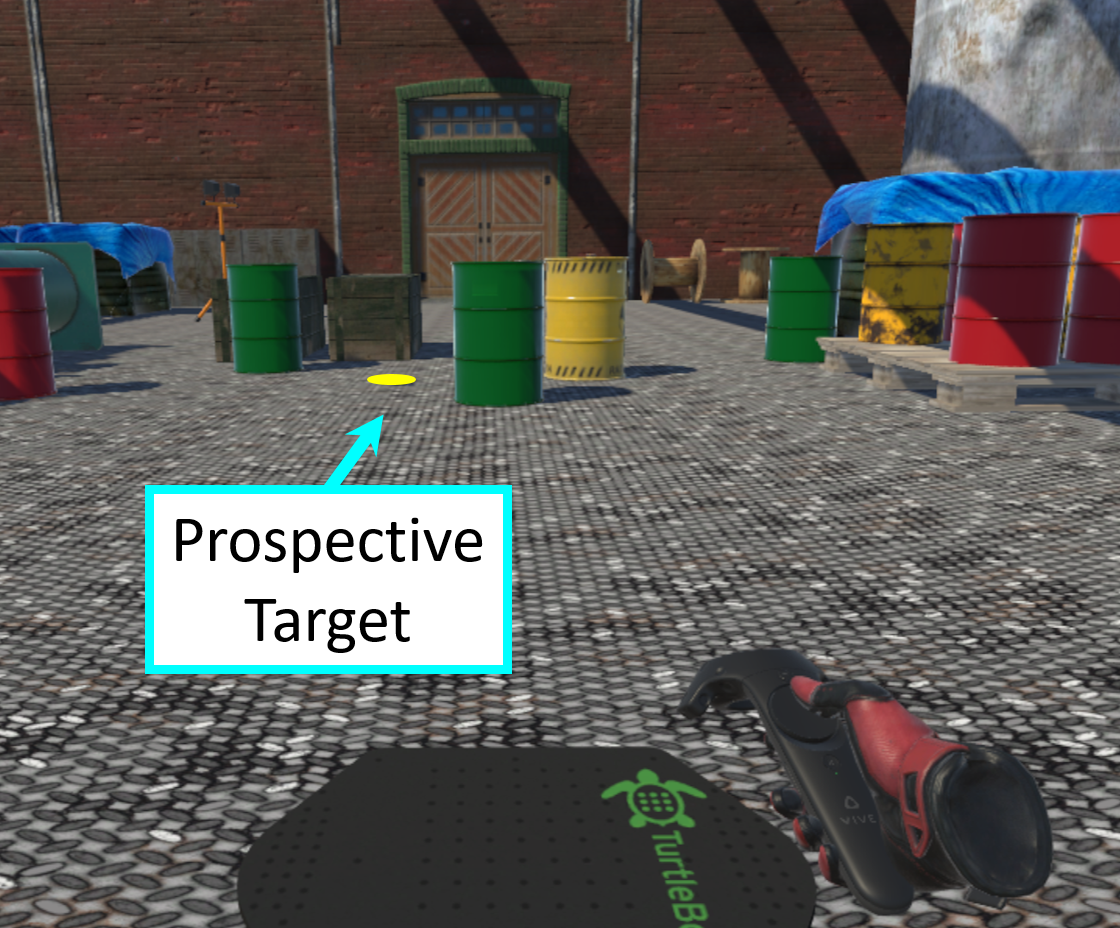} 
    \caption{Prospective target displayed}
    \label{fig:prospective_target}
  \end{subfigure}
  \hfill
  \begin{subfigure}[b]{0.32\hsize}
    \centering
    \includegraphics[width=\hsize]{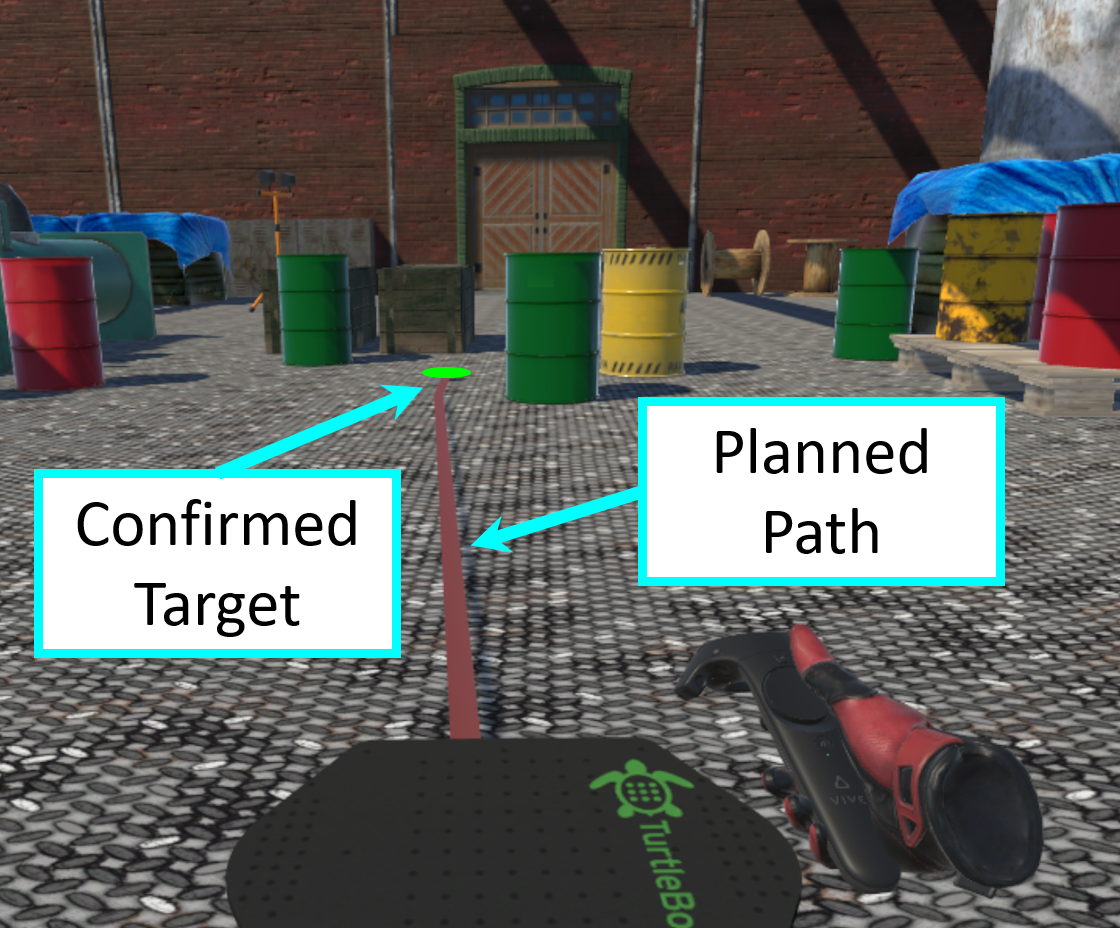} 
    \caption{Robot moves to target autonomously}
    \label{fig:confirmed_target}
  \end{subfigure}
  \caption{The principle steps of the waypoint control method: (a) target position is selected using the tracked VR controller; (b) target is displayed and confirmed by the user; (c) the robot moves autonomously to the target and publishes its path to the operator.} 
  \label{fig:waypoint_control}
\end{figure*}

The proposed target selection strategy, is based on a method of locomotion called \textit{teleportation} used in the VR gaming industry. Teleportation was designed to allow players to navigate around virtual environments in an intuitive manner. To perform teleportation, the player selects a point in the environment using the 6DoF tracked VR controllers as a virtual pointing device. On selection of a point, the player's position in the environment is instantly changed to the chosen position. While it is clearly not possible to teleport a robot between locations, this method does seem to be a convenient and intuitive way to select target points in VR.

The target selection we propose uses the same input method as teleportation. However, instead of transporting the user directly to the target, the position is logged and used as a goal location for the robot’s onboard planner. The proposed method is shown in \textbf{Figure \ref{fig:waypoint_control}}, and in the accompanying video \footnote{\url{https://youtu.be/yW6j36XRT_8}}. The input method works as follows: (1) The operator depresses a button on the VR controller, which causes a virtual arc to render between the controller and the ground plane (\textbf{Figure \ref{fig:target_select}}). The operator can alter the position of the target reticle by changing the position and orientation of the controllers, thereby changing the point at which the arc intersects the terrain. (2) When the operator is satisfied with the position of the target reticle, they release the button and a yellow prospective target disk is spawned at that location (\textbf{Figure \ref{fig:prospective_target}}). (3) The operator then confirms the selected target by pulling the trigger on the controller. The target disk changes colour to green and the robot starts planning a path to the chosen target position. Once a path has been calculated, it is communicated back to the operator, and superimposed on their display (\textbf{Figure  \ref{fig:confirmed_target}}). (4) While the robot is moving to the target, the operator may select a new target position by repeating steps (1) - (3), or can order the robot to stop moving immediately using the grip button on the controller.

One of the benefits of this target selection method is that it can work across a range of HMD-based teleoperation setups, using a standard commercial VR device (e.g. HTC Vive). \citet{Roldan2019BringingInfrastructures} showed that the target selection method is effective in a hand-modelled virtual environment, for a multi-robot use case. We also suggest that this method could be used for a collocated operator and robot use case (e.g. \cite{Walker2019Robot}), by superimposing the target-setting virtual arc onto the real environment using an ARHMD (e.g. Microsoft HoloLens). Similarly, this target-selection method is also possible when viewing the camera feed from the remote robot directly (i.e. an egocentric viewpoint), by superimposing the virtual arc onto the camera feed. This is the type of interface we are interested in developing, because it does not necessitate detailed 3D mapping of the environment.

As a first step towards this type of interface, we developed a VR testbed in which users can control a virtual robot around a virtual environment using the aforementioned target selection method. This allowed us to examine various interface design decisions in preliminary trials, ultimately culminating in a usability study on the proposed method, outlined in \textbf{Section \ref{sec:experiment}}. Such experiments are important before implementation on a real robotic system, as they help ensure that the interface is viable for operators to use and provide quantitative data on which to justify interface design decisions. This should help maximise the effectiveness of the future interface and reduce the negative effects, such as minimising the VR sickness that operators feel.

\section{Methods} \label{sec:experiment}

We conducted a user study to test the effectiveness of the proposed target selection method when using an egocentric viewpoint. The primary objective of the experiment was to navigate a virtual TurtleBot 2 robot from one end of an indoor environment, to a goal position at the far end. The aim of the study was to provide quantitative data and qualitative feedback on the effectiveness and usability of the interface. The experiment was run entirely in virtual reality; i.e. the robot controlled in the experiment was not a virtual ‘twin’ of a physical system operating in a real-world environment. The experiment was conducted using an HTC Vive HMD, with standard 6DoF tracked Vive controllers. The virtual environment, the robot, and the integration with the HTC Vive, were implemented using the Unity game engine \footnote{\url{https://unity.com}}. In this section we will first provide details of the experimental design, including the apparatus used and a description of the test environment. We also provide an outline of the experimental procedure, and identify the measures recorded during the experiment.

\subsection{Experimental design}

The waypoint navigation interface, with the proposed target selection method, was compared against a baseline direct control method. In the direct control condition, the operator used the 2-axis trackpad on the Vive controller to provide control inputs. The position of the operators thumb on the y-axis of the trackpad was used to control the forward/back motion of the robot and the position on the x-axis was converted to left/right turning motions. In this condition, a collision avoidance mechanism was implemented that prevented the robot from driving into obstacles in the environment; a function which is common in teleoperated systems. A UI warning was displayed if the proximity sensor on front of the robot was triggered and the robot was prevented from moving forward, and instead restricted to turning or reversing motions. In both setups, the operator had a full 360\degree view of the environment, viewable by turning their head in the VR headset. However, unlike standard VR-gaming interfaces, the spatial position of the operators viewpoint was fixed to the position of the camera on the robot, to better represent the constraints of a real-world camera system. The VR system also renders a stereoscopic pair of images to the operator each frame, allowing them to perceive depth more easily than a monocular video. 

As discussed in \textbf{Section \ref{sec:background}}, communication delays between the operator and the robot are often present in real-world systems. One of the simplifications of running a fully virtual experiment is that these communication delays are essentially negated. In order to account for this, we included an artificial communication delay in the system as a variable. In the experiment, we tested two delay conditions; no delay and a fixed 1 second delay. A fixed delay was chosen in order to minimise the possible VR sickness effects that the participant would experience. The 1 second duration of the delay was chosen because it is in the range at which direct control performance becomes significantly degraded \cite{Chen2007HumanRobots} and aligns with the ‘short’ time delay duration used by \citet{Luck2006AnLatency} for a mobile robot navigation task.

A 2x2 within-subjects factorial design allowed us to test the effectiveness of the proposed interface against direct control, across two levels of delay (no delay and 1 second delay). The within-subject design allowed us to gather direct comparative feedback from the participants. The order of the control conditions was counterbalanced. However, the non-delayed condition was always run first for a given control method. This ordering was chosen so that the participant did not have to experience an unfamiliar control method and adverse delay conditions simultaneously, as it was thought that this may be too stressful and exacerbate the effects of VR sickness. 

\subsection{Experimental Procedure}

During the experiment, the participant experienced the following set of events:
\begin{enumerate}
    \item A short video outlining the goal of the experiment \footnote{\url{https://youtu.be/oJFDSp7mYjw}}.
    \item A VR-based tutorial for the first control method they were going to use (direct control or waypoint navigation). During the tutorial, the participant was able to try controlling the robot with no delay and 1 second delay.
    \item A timed navigation trial through the virtual environment, with no communication delay. The participant then answered questionnaires about the trial (see Measures).
    \item Another timed trial, with the same control method, but with 1 second communication delay, followed by questionnaires.
    \item A repeat of Steps (2) - (4), using the other method of control.
    \item An optional ‘bonus’ trial, where the operator could switch between the two control methods as and when they chose (described below).
    \item A semi-structured discussion about the experiment (see Measures).
\end{enumerate}

\begin{figure*}
    \centering
    \includegraphics[width=\textwidth]{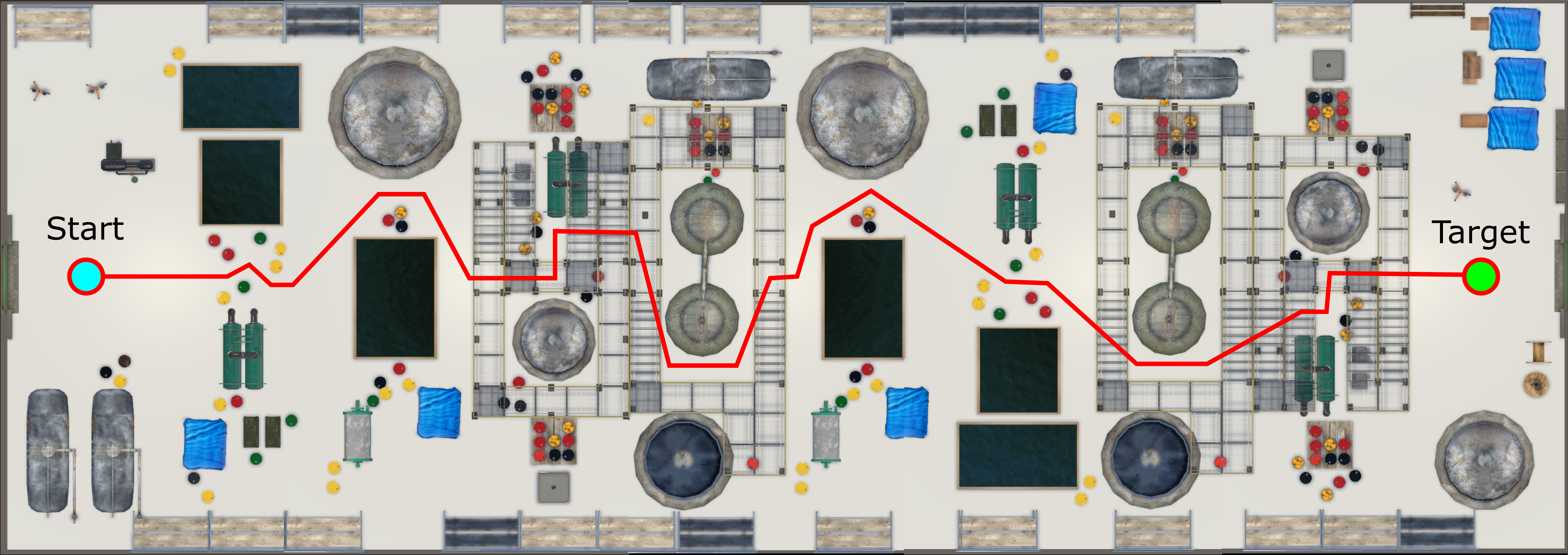} 
    \caption{Top down view of the VR trial environment, overlaid with the optimum route between the start and finish points.} 
    \label{fig:environment}
\end{figure*}

A top down view of the virtual environment used in the timed trials is shown in \textbf{Figure \ref{fig:environment}}. The environment contained a significant number of static obstacles, including barrels, waste silos, and pillars, which supported elevated walkways. The environment was used in the forward direction (Start to Target in \textbf{Figure \ref{fig:environment}}) for the first trial (without delay), and in reverse (Target to Start) for the second trial (with delay) in order to keep the optimal path length the same. Mirrored versions of these two environments were used in the third and fourth trials (forward-mirrored and reverse-mirrored respectively) to ensure that the environment was identical between control methods for a given delay level. The potential for learning effects from reusing the same environment (albeit mirrored) was counteracted by counterbalancing the order of the control conditions. In piloting, very few participants realised that the same environment layout was being used, so learning effects due to this are likely to be minimal in any case.

In order to gauge the participant’s cognitive workload during the navigation task,  the participants were also asked to perform a timed ancillary task. Left or right facing arrows would periodically appear ($\approx 8$ seconds apart) on either side of their screen, and the participant was required to click the corresponding button on the VR controller in their non-dominant hand; the dominant hand controller was being used to manoeuvre the robot. The response times to the stimuli were measured throughout the timed trials.

After the four timed trials, participants were given the option to partake in an additional `bonus' round. The aim of the bonus round was to garner objective data about the participant’s preferred control method in different situations. In the bonus round, the participant was given the ability to switch between the two control methods (direct control and waypoint navigation). In the first half of the trial, no communication delay was present, but halfway through the trial a 1 second delay was imposed; the participant was notified before the trial that this was going to happen, and was alerted during the trial when it occurred. It was made known to the participants that a small monetary prize would be awarded to the participant with the lowest overall time (the sum of trial completion time and the cumulative response times to the ancillary task), in order to encourage them to optimise their task speed. In real-world mobile robot navigation missions, task speed is only one of several factors that operators may seek to optimise, but nevertheless provides a useful basis for comparison in this experiment.

\subsection{Measures}

Our overall aim was to develop an effective and intuitive target selection method for HMD-based teleoperation. We decided to compare the effect of the proposed interface based on three factors: (a) task performance (b) cognitive workload, and (c) usability. We used both objective and subjective measures to evaluate the effects of the interface on these areas. Task performance and cognitive workload comparisons between the novel control method (waypoint control) and the familiar control method (direct control) in the non-delayed condition were chosen to show whether participants were able to pick up the novel interface quickly, thereby indicating its intuitiveness.

For task performance, primary (navigation) task completion time and total distance travelled by the robot, were used as objective measures.

For cognitive workload, response times to the ancillary (button-pressing) task and the number of ancillary task errors (i.e. pressing the wrong button) were used as objective measures. As a subjective measure, we decided to use the `Raw' variant of the NASA task load index (Raw-TLX) questionnaire, where the final score is simply taken as a mean of the scores from each category: mental demand, physical demand, temporal demand, performance, effort and frustration. This method has been shown to be equally as effective as the original weighted NASA-TLX in a number of studies \citep{Hart2006NASA-taskLater}, and takes significantly less time to administer.

As a subjective measure for system usability, we utilised the system usability scale (SUS) questionnaire \cite{Brooke1996SUS-AScale} after each trial. Qualitative feedback was obtained from the participants using a semi-structured interview at the end of the trial. During the interview participants were asked to state their preferred control method in the case of no delay, and in the case of a delay. Participants were also asked open-ended questions, such as "what were your thoughts on the two methods of control you used in this experiment?". The amount of time each control method was used in the bonus trial was also recorded as an objective measure of user preference.

\section{Results} \label{sec:results}

\subsection{Participants}

A total of 37 participants were recruited from the university campus. 3 participants (1 male, 2 female) did not finish all 4 trials, because they experienced significant nausea effects. Data from their trials was removed, leaving 34 participants (25 male, 9 female). Demographics recorded for the participants included how often they played computer games (2 daily, 16 weekly, 12 monthly, 4 never), experience with virtual reality (1 a lot, 29 some, 4 none) and experience operating robots (12 a lot, 22 some). The average participant age was 28.2 ($\sigma$ = 4.46), with a range of 22-43. Out of the 34 participants that completed all 4 conditions, 22 chose to participate in the bonus round.

\subsection{Analysis}

Unless otherwise stated, the measures outlined above were analysed using a repeated-measures two-way analysis of variance (ANOVA) with control method and delay as the two independent variables, each with two levels. The order that participants experienced the two control conditions was treated as a between-subject factor, to analyse the effect that order had on the results. The error bars in the following graphs show the 95\% confidence interval.

\subsection{Objective Measures}

\begin{figure*}
\centering
  \begin{subfigure}[b]{0.32\hsize}
    \centering
    \includegraphics[width=\hsize]{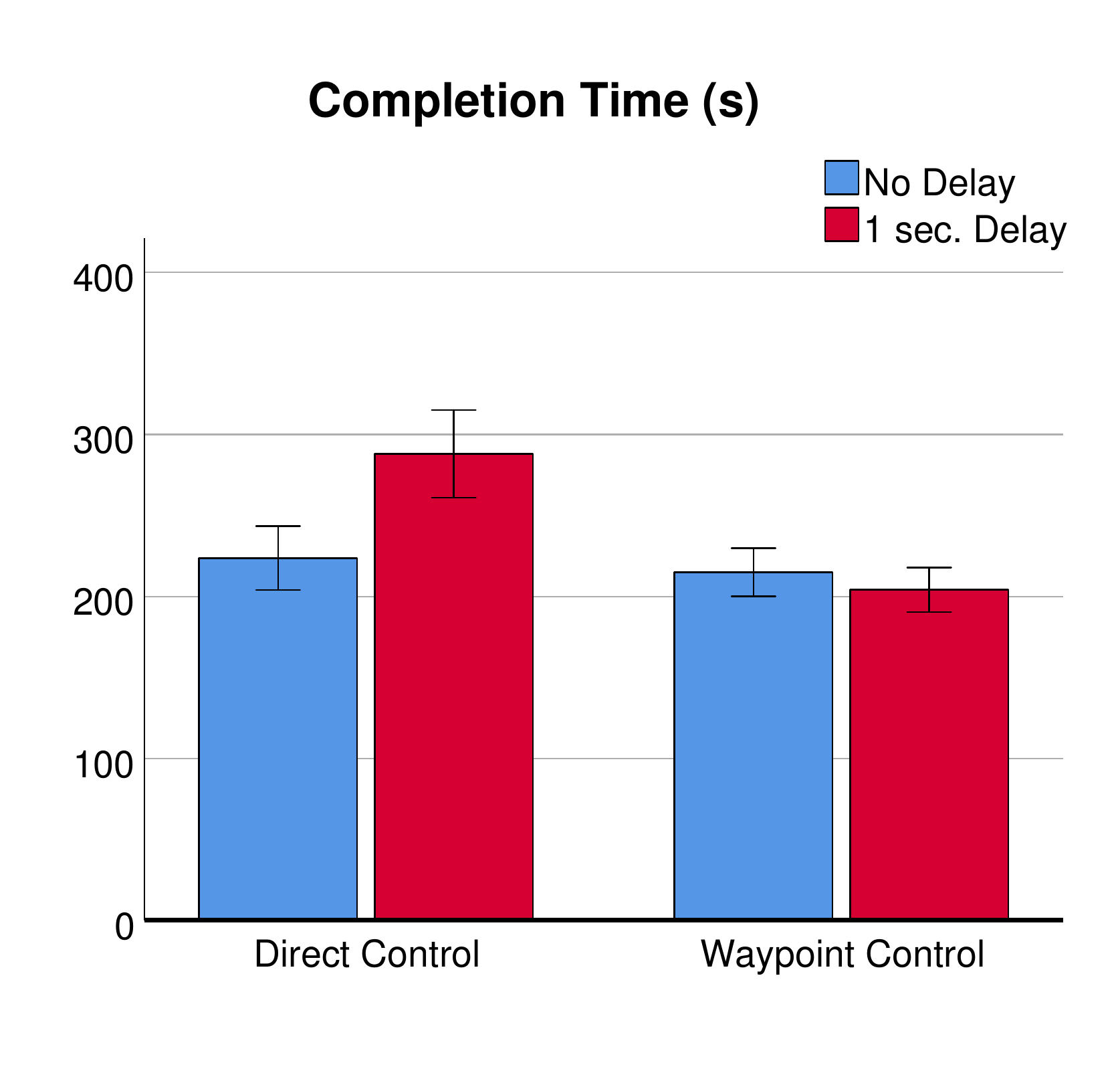}
    \label{fig:completion_time}
  \end{subfigure}
  \vspace{-1cm}
  \begin{subfigure}[b]{0.32\hsize}
    \centering
    \includegraphics[width=\hsize]{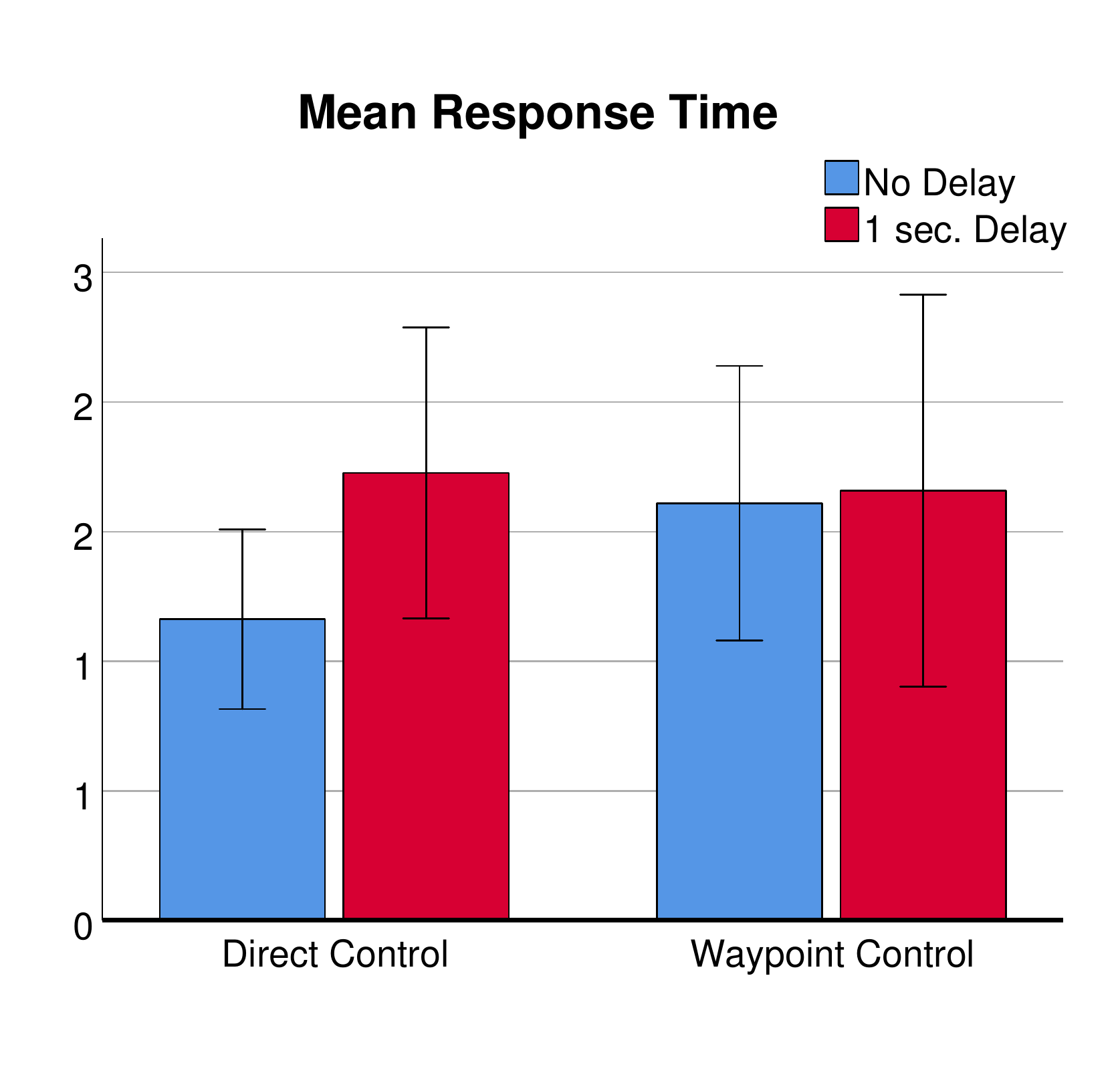}
    \label{fig:response_time}
  \end{subfigure}
  \begin{subfigure}[b]{0.32\hsize}
    \centering
    \includegraphics[width=\hsize]{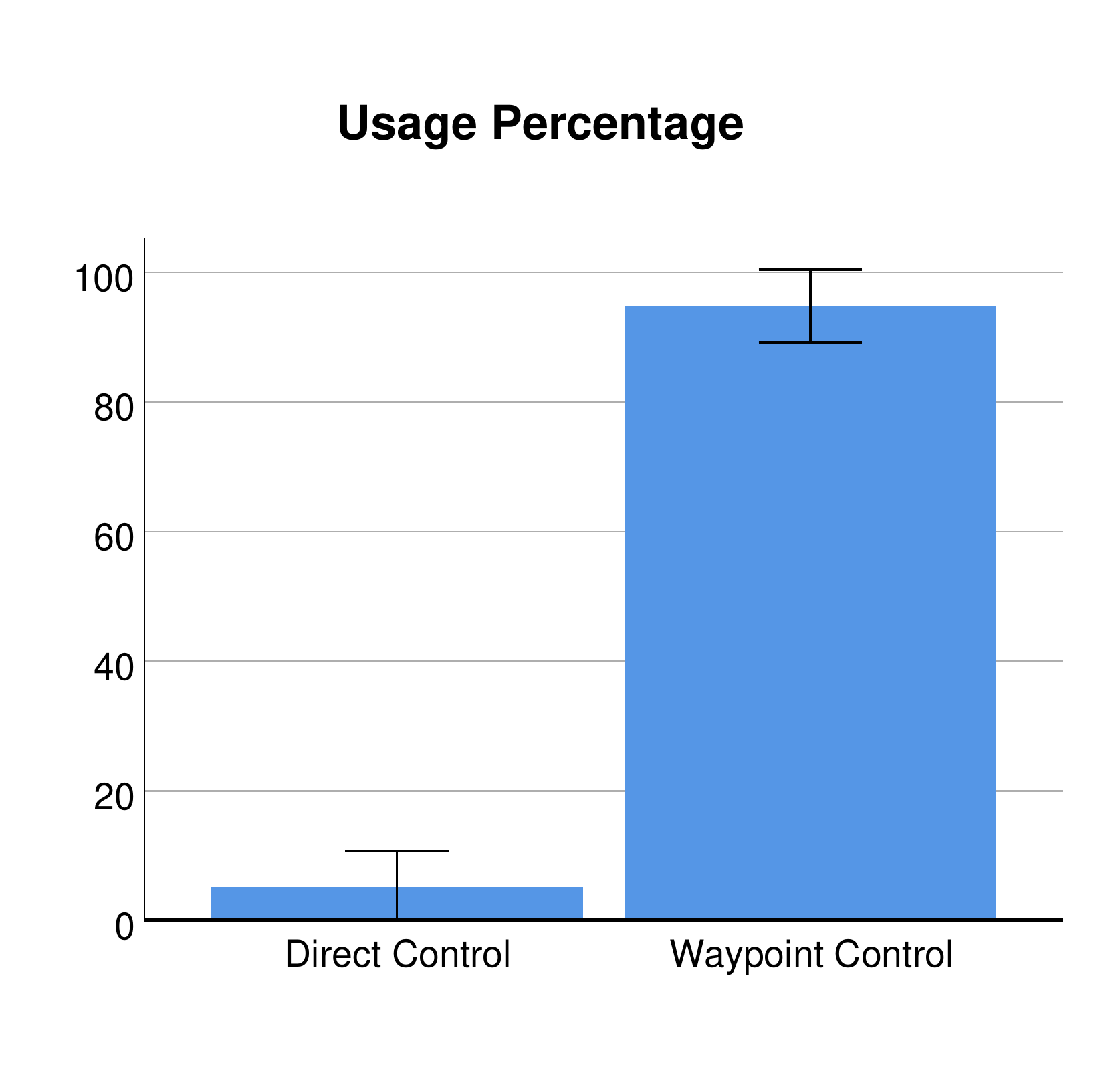}
    \label{fig:usage}
  \end{subfigure}
  \caption{Results of objective measures. Error bars at 95\% confidence intervals.} 
  \label{fig:objective_measures_results}
\end{figure*}

The repeated-measures two-way ANOVA showed a significant interaction effect between control method and delay on task completion time, $F(1,32) = 28.86, p < 0.001$. Order did not have a significant effect on this measure. Simple effects analysis showed that there was no significant difference between the control methods when delay was not present ($p = 0.2$). However, in the delayed condition waypoint control significantly outperformed direct control, $F(1,32) = 57.39, p < 0.001$. Analysis showed that there were no significant main effects of control or delay on mean response time to the secondary task. 
The usage percentages of each control condition during the bonus trial, where participants could switch between control methods at any time during the trial, were analysed using a paired samples t-test. The results showed a significant usage preference for waypoint control ($M = 94.8\%$) over direct control (M = 5.2\%), $t(22) < 0.001$.

\subsection{Subjective Measures}

\begin{figure*}
  \begin{subfigure}[b]{0.32\hsize}
    \centering
    \includegraphics[width=\hsize]{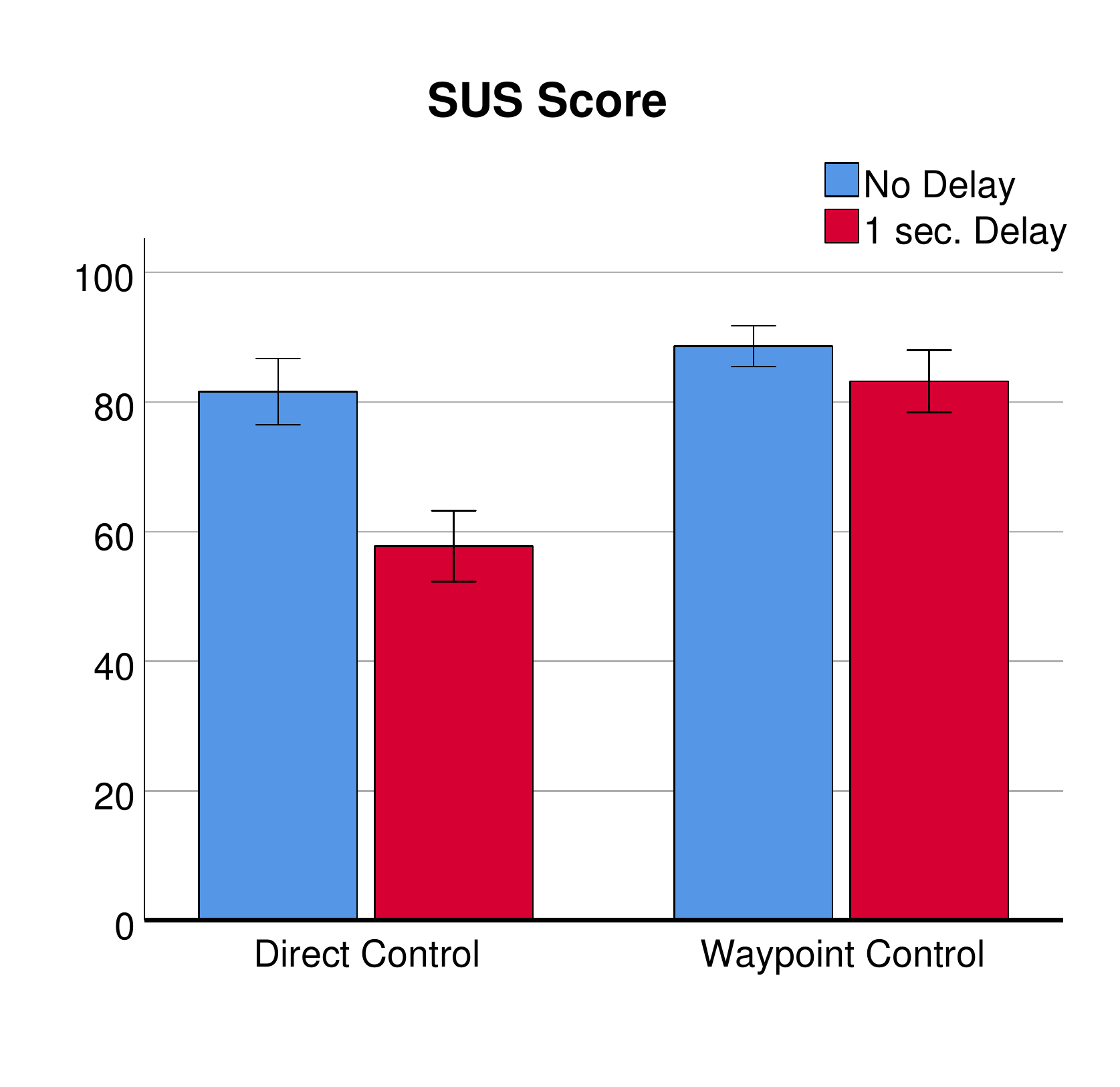}
    \label{fig:sus_results}
  \end{subfigure}
  \begin{subfigure}[b]{0.32\hsize}
    \centering
    \includegraphics[width=\hsize]{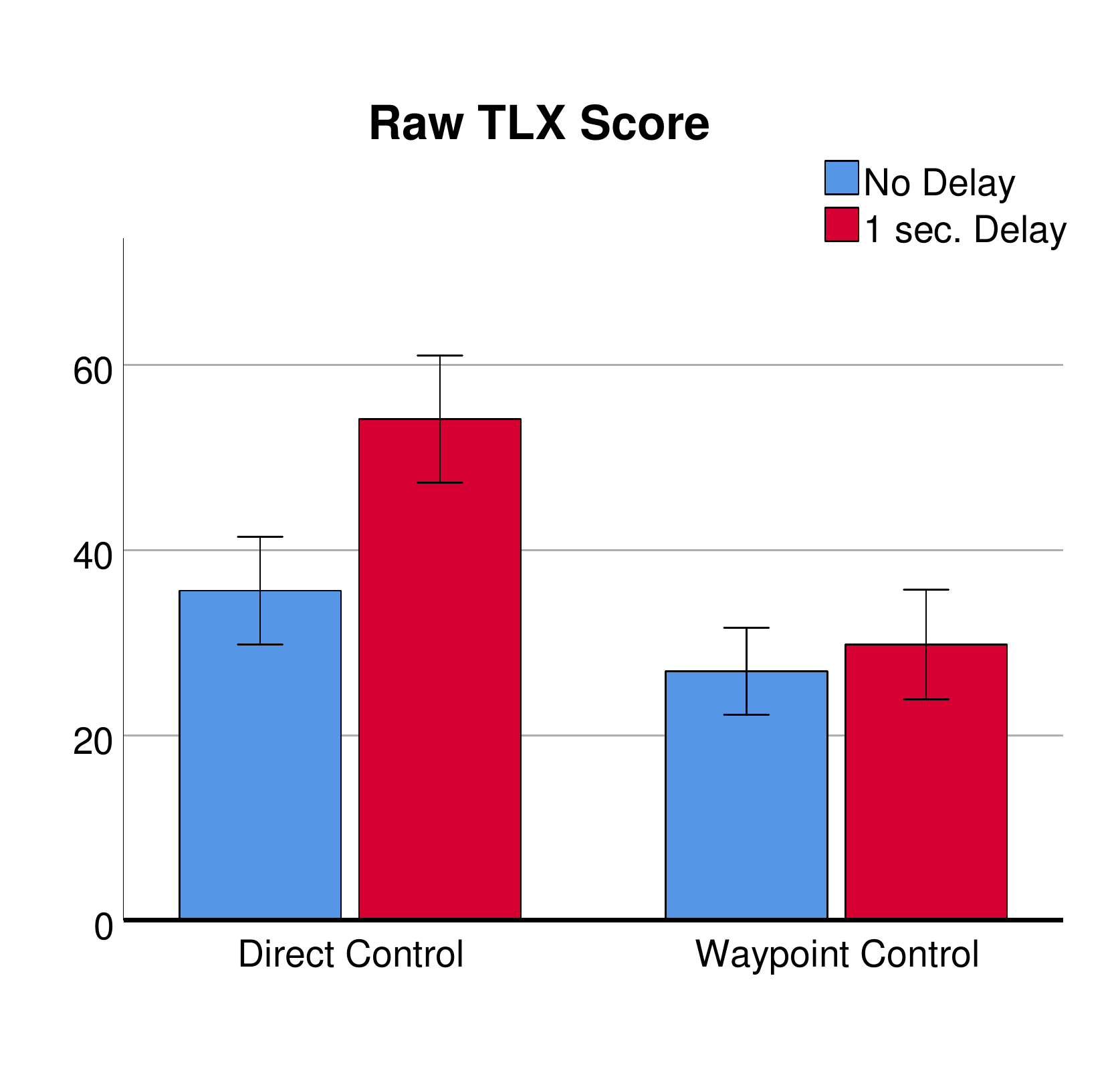}
    \label{fig:tlx_results}
  \end{subfigure}
  \vspace{-1cm}
  \caption{Results of subjective measures. Error bars at 95\% confidence intervals.} 
  \label{fig:subjective_measures_results}
\end{figure*}

Control interface usability was evaluated with the SUS questionnaire. A two-way repeated-measures ANOVA showed a significant interaction between control method and delay on the total SUS score, $F(1,32) = 25.889, p < 0.001$. The analysis also showed a significant interaction between control method and order, $F(1,32) = 7.795, p = 0.009$. \textbf{Table \ref{tab:sus_split}} shows the estimated marginal means of the SUS scores when split by order; direct control first (DC First) and waypoint control first (WC First). For the group that experienced the direct control condition first, there was no significant difference between the SUS scores in the non-delayed conditions ($F(1,18) = 0.25, p = 0.877$). However, for the group that experienced waypoint control first, the SUS scores for waypoint control were found to be significantly higher, $F(1,14) = 7.545, p = 0.016$. For both groups, the SUS scores were found to be significantly higher for waypoint control in the delayed case ($F(1,18) = 54.724, p < 0.001$ and $F(1,14) = 36.977, p<0.001$). When the data from both groups is combined into one set, simple effects analysis showed a significantly higher total SUS score for waypoint control in both the non-delayed condition, $F(1,32) = 7.21, p = 0.011$, and the delayed condition, $F(1,32) = 88.89, p < 0.001$.

\begin{table}[]
\caption{Estimated marginal means ($M$) and standard deviations ($\sigma$) of SUS scores when the data is split into 2 groups by order of control conditions; direct control (DC) first and waypoint control (WC) first.}
\label{tab:sus_split}
\begin{tabular}{cc|cc|cc|}
& &  \multicolumn{2}{c|}{DC First} & \multicolumn{2}{c|}{WC First}\\
Control & Delay & $M$ & $\sigma$ & $M$ & $\sigma$ \\
\midrule
\multirow{2}{*}{DC} & None & 86.81 & 2.01 & 77.00 & 5.05 \\
& 1 sec. & 62.37 & 2.67 & 53.17 & 5.02 \\
\midrule
\multirow{2}{*}{WC} & None & 86.58 & 2.34 & 90.67 & 1.80 \\
& 1 sec. & 83.55 & 3.10 & 82.83 & 3.59 \\
\end{tabular}
\end{table}

The Raw-TLX was used to assess cognitive workload during the tasks. A two-way repeated-measures ANOVA showed a significant interaction between control method and delay on the total Raw-TLX score, F(1,32) = 19.182, p < 0.001. A simple effects analysis showed a significantly lower Raw-TLX score for waypoint control in both the non delayed condition, F(1,32) = 8.713, p = 0.006, and the delayed condition, F(1,32) = 66.472, p < 0.001.

\textbf{Table \ref{tab:raw_tlx_stats}} shows the significance values for the difference between waypoint and direct control on the components of the Raw-TLX. These results were obtained using a two-way repeated-measures ANOVAs followed by post hoc simple effects analyses.

\begin{table}[]
\caption{Statistical significance test results for Raw-TLX component scores. In all significant cases, waypoint control condition had lower mean values than direct control.}
\label{tab:raw_tlx_stats}
\begin{tabular}{lll}
\toprule
Component          & No Delay                  & 1 Second Delay                 \\
\midrule
Mental & Not Significant           & F(1,32) = 29.01, p \textless 0.001\\
Physical & F(1,32) = 7.62, p = 0.009 & F(1,32) = 14.42, p =0.001   \\
Temporal & F(1,32) = 5.90, p=0.021   & F(1,32) = 28.11, p \textless 0.001 \\
Performance     & Not Significant           & F(1,32) = 15.91, p \textless 0.001 \\
Effort          & F(1,32) = 6.53, p = 0.016 & F(1,32) = 56.91, p \textless 0.001 \\
Frustration     & F(1,32) = 7.14, p = 0.12  & F(1,32) = 50.33, p \textless 0.001\\
\midrule
\end{tabular}
\end{table}

Finally, participants were asked to state a preferred control method in a delayed and non-delayed condition. In the non-delayed condition, 2 participants (6\%) stated a preference for direct control, 4 participants (12\%) stated no preference, and the remaining 28 participants (82\%) stated a preference for waypoint control. In the delayed condition, all participants stated a preference for waypoint control.

\section{Discussion} \label{sec:discussion}

The aim of this work is to develop an effective and intuitive target selection method for HMD-based teleoperation. The results show that the waypoint control interface used in this experiment was either similar or better than direct control in all three areas of interest (task performance, cognitive workload and usability) across both levels of delay (no delay and 1 second delay). This result comes in spite of the fact that direct control is a more familiar method of control to most users; highlighting that participants were able to pick up the target selection method quickly and thus its intuitive nature.

One overwhelming result, observed across all recorded metrics, was that the waypoint control method was significantly more effective than direct control when time delay was present. Many participants fed back that the waypoint control method was barely affected by the delay, and some (albeit few) actually preferred it with delay, stating that they felt it gave them more time to think. This result shows that our proposed interface successfully leverages one of the primary benefits of supervisory control schemes; robustness in the face of communication delay.

The results of the control method rankings and the usage percentages in the bonus round show that participants preferred using the waypoint control system, even when there was no delay, for this task. Comments from participants suggest that transferring the responsibility for local path-planning and collision avoidance to the robot allowed them to concentrate on higher level route planning. This was also reflected in the higher usability scores of the waypoint control interface, particularly in the delayed condition. Order effects were shown to be significant on usability scores; participants scored the usability of direct control significantly lower when they experienced it after waypoint control; i.e. participants felt that usability was impacted more when autonomous functionality was removed, than when it was added, which is an interesting result.

The Raw-TLX scores indicated significantly lower cognitive load when using waypoint control. However, response time to the ancillary task (the objective measure for cognitive load) did not show the same trend. In fact, in the conditions with no delay, mean response times to the secondary task appears to be lower in direct control than waypoint control (albeit not significantly). Based on observations in the trials and participant comments, the immediacy and simplicity of the secondary task meant that it was not severely affected by their cognitive workload levels. Furthermore, experimental observations indicated that waypoint control induced more fluctuation in the operators cognitive load due to the primary task. A high level of cognitive load was required while the user was selecting a target position, followed by a period of much lower cognitive load while the robot was moving autonomously. Because the secondary task required an immediate response, the results could have been affected by these short periods of high workload. Additionally, reduced focus during the autonomous manoeuvres may have also negatively affected response time. A more complex ancillary task may have provided better granularity on these cognitive workload fluctuations, and may have highlighted the differences between the two control methods more clearly. An entirely different ancillary task (e.g. being required to search for and count particular objects in the environment) might have better highlighted the effect of the participants’ cognitive workload on their situational awareness, which was not explicitly measured in this experiment. Subjective workload results displayed the same ordering effects as usability, we expect for the same reasons.

VR sickness turned out to be a significant factor in this experiment, with 3 participants unable to finish all 4 conditions, and a significant proportion of participants reporting mild to moderate nausea in the post experiment interviews. Unfortunately, nausea effects were not systematically recorded as a measure in this experiment, but participant comments highlighted that there could be an impact of control method on levels of nausea experienced. The direct control condition with delay was most frequently cited as the condition that induced the most nausea. This could be in part because of the higher level of workload and concentration required during this condition. Overall, VR sickness effects are likely to have a large impact on whether egocentric HMD-based displays are viable for mobile robot teleoperation and their effects should be measured in future experiments.

\subsection{Limitations and Further Work}

We made the decision to run the experiment virtually in order to get an indication of how an optimal implementation of this type of interface might perform. However, several additional challenges need to be tackled when applying this to a real-world system. The robot must be able to map its environment in great enough resolution for the operator to accurately select a target using the proposed method. However, because the operator uses camera images for scene understanding when using an egocentric viewpoint, the level of detail required of the 3D model should still be significantly less than that required for an exocentric implementation, where the 3D model is the operators only way of viewing the environment.

The actual communication delay present in a real-world implementation could also feasibly be shorter or longer than 1 second, and/or variable (rather than fixed), depending on factors associated with a specific setup (range of teleoperation, communication methods, camera type etc.).

When implemented on a real robot, we also intend to compare this target selection method to others, such as using a 3D joystick to place targets (e.g. \cite{Cameron1987FusingVehicles}). We also intend to compare the effectiveness of egocentric and exocentric viewpoints. Finally, in future experiments we aim to measure nausea effects more systematically (e.g. using the Simulator Sickness Questionnaire \cite{Kennedy1993SimulatorSickness}), providing insight into which types of interface cause the least VR sickness effects.

\section{Conclusion} \label{sec:conclusion}

This paper has explored the utility of an egocentric HMD-based target selection interface for waypoint navigation of mobile robots. The proposed target selection method was based on the VR gaming notion of \textit{teleportation}. We implemented this method using an HTC vive and accompanying tracked controllers to select waypoints in a virtual reality environment.

Both the effectiveness and the intuitiveness of the proposed interface were examined through a user study, involving 37 participants. The results of the VR-based user study showed that this type of target selection method is viable for implementation on real robots. Participants were able to learn how to use the system quickly, and were therefore able to leverage the beneficial characteristics of semi-autonomous control methods; namely reduced cognitive workload and mitigated communication delay effects. The experiment also highlighted that an egocentric viewpoint may be problematic for HMD-based teleoperation, because the motion of the robot's camera through the environment, viewed through the VR headset, caused significant nausea effects in some participants.

Further work will involve implementing this target selection method on a real robot, and comparing it to other methods of target selection currently used for mobile robots.

\bibliographystyle{ACM-Reference-Format}
\bibliography{references}

\end{document}